\documentclass{article}
\usepackage{spconf,amsmath,graphicx}
\usepackage{booktabs}
\usepackage{xcolor}

\title{Spatial Moment Pooling Improves Neural Image Assessment}
\name{Tongda Xu$^{\star}$ \qquad Yifan Shao$^{\dagger \star 1}$\thanks{1. This work is done when Yifan is an intern with Sensetime Research.} \qquad Yan Wang$^{\ddagger \star 2}$\thanks{2. To whom correspondence should be addressed.} \qquad Hongwei Qin$^{\star}$}
  
\address{$^{\star}$ Sensetime Research \qquad $^{\dagger}$ Beihang University \qquad $^\ddagger$ Tsinghua University}

\begin{document}
%
\maketitle
\begin{abstract}
In recent years, there has been widespread attention drawn to convolutional neural network (CNN) based blind image quality assessment (IQA). A large number of works start by extracting deep features from CNN. Then, those features are processed through spatial average pooling (SAP) and fully connected layers to predict quality. 
Inspired by full reference IQA and texture features, in this paper, we extend SAP ($1^{st}$ moment) into spatial moment pooling (SMP) by incorporating higher order moments (such as variance, skewness). Moreover, we provide learning friendly normalization to circumvent numerical issue when computing gradients of higher moments. Experimental results suggest that simply upgrading 
SAP to SMP significantly enhances CNN-based blind IQA methods and achieves state of the art performance.

\end{abstract}
\section{Introduction}
\label{sec:intro}

Different from full reference image quality assessment (IQA) (e.g. PSNR, SSIM), blind IQA computes quality directly from the distorted image without a clean image, which makes it suitable to user generated content, communication and other scenarios where clean sources are inaccessible. Traditional blind IQA approach relies on the regularity of natural scene statistics. Hand-crafted features \cite{mittal2012no} are designed to measure the distortion of image from natural ones.

More recently, deep CNN based blind IQA approaches have shown promising results as learned features are more powerful \cite{kang2014convolutional}. And a typical inference of such IQA approach involves several steps: (1). a CNN backbone (e.g. VGG16) is applied to extract deep features from distorted images. (2). the deep features are spatially average pooled (SAP). (3). the averaged features are passed through a regression head (e.g. fully connected layers) to predict image quality. 
In some works, the CNN backbone is pretrained on other tasks (e.g. ImageNet) and finetuned with regression head \cite{talebi2018nima}\cite{su2020blindly}\cite{chen2021nested}. While for other works auxiliary tasks such as predicting residual or noise type are added \cite{kim2018deep}\cite{chen2021nested}. Despite variation in the training methods, the inference procedures are similar.

\begin{figure}[thb]

\includegraphics[width=8.5cm]{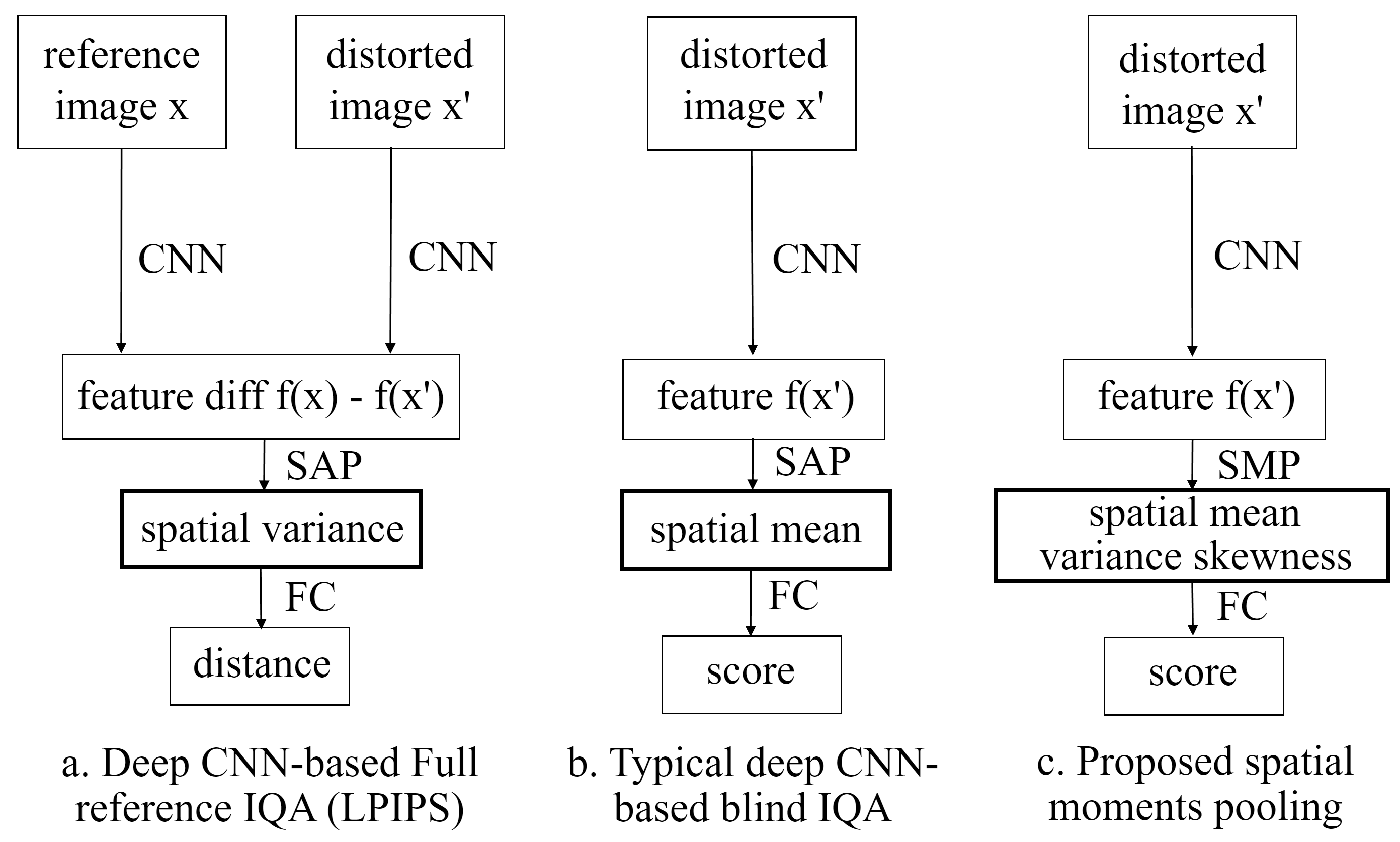}
\label{fig:res}
\caption{(a). CNN based full reference IQA, (b). CNN based blind IQA and (c). our proposed SMP approach.}

\end{figure}

Formally, denote distorted image as $x' \in R^{H \times W}$, the feature from CNN as $f(x) \in R^{C \times H' \times W'}$, then the output of spatially average pooling $SAP(f(x'))$ can be seen as spatial mean of feature map. And the pooled feature map is further processed by head $g$ to obtain the predicted quality $d(x')$.

\begin{equation}
d(x') = g(SAP(f(x'))) = \\ g(E[f(x')])
\label{eq:1}
\end{equation}

On the other hand, CNN based full-reference IQA methods also show promising results. As a representative method, LPIPS \cite{zhang2018perceptual} has been widely adopted as perceptual loss \cite{mentzer2020high}. Similar to blind IQA, LPIPS first computes the feature maps of reference image $x$ and distorted image $x'$ from a CNN $f$. Then the square difference of those feature maps are computed and spatially averaged. Finally, the averaged results are processed by head $g$ to predict the distance metric $d(x, x')$.

\begin{equation}
  \begin{array}{l}
    d(x, x') = g(SAP((f(x) - f(x'))^2)) \\ \hspace{3.5em} = g(Var[f(x) - f(x')])
  \end{array}
\label{eq:2}
\end{equation}

Though both Eq.~\ref{eq:1} and Eq.~\ref{eq:2} adopt SAP over features, the statistics computed are different (See Fig. 1). For blind IQA, the SAP is used to compute $1^{st}$ moment (mean) of feature map. For LPIPS, by assuming the residual has $0$ mean, the SAP computes the $2^{nd}$ central moment (variance) of feature map difference. 

For long, variance has been an important part of texture feature \cite{haralick1979statistical,sajjadi2017enhancenet}. Higher moments (e.g. skewness, kurtosis) are also applied in texture analysis \cite{shaban2001improvement}. So a natural question to ask is, is spatial variance and other moments more representative than mean alone for image quality?

In this paper, we propose spatial moment pooling (SMP) as an extension of SAP for blind IQA. Our approach of adding higher moments absorbs the arts of full reference IQA methods and texture features. To be specific, for each channel of feature map, those higher central moments (e.g. variance) are also computed and concatenated with $1^{st}$ moment. It can be seen as a generalization of SAP ($1^{st}$ moment). Moreover, we also propose a normalization approach to avoid numeric issue when back-propagating through higher order moments. Experimental results show that by simply replacing SAP with SMP, the performance of many deep CNN based IQA methods is significantly improved. Our method outperforms previous CNN approaches.

\section{Related Works}
\label{sec:rw}

\subsection{Poolings and Deep CNN based blind IQA}

Despite the pioneer of CNN based blind IQA \cite{kang2014convolutional} adopts min-max-pooling, the majority of subsequent works \cite{su2020blindly}\cite{chen2021nested}\cite{chadha2021deep}\cite{kim2016fully} adopts SAP (include global average pooling) as it has become popular in CNN. \cite{ying2020patches} adopts a mixture of SAP and ROI Pooling. \cite{talebi2018nima} and \cite{ke2021musiq} resize input images or patches to fixed size instead, and do not use pooling for final features.

On the other hand, covariance pooling \cite{wang2020deep} has been shown effective for high level vision tasks. And the $i = 2$ moment is the diagonal elements of a covariance matrix. However, to the best of our knowledge, neither moment pooling with $i \ge 3$ nor moment pooling with $i \ge 2$ for CNN based blind IQA has been studied.

\section{spatial Moment Pooling}

\subsection{Background}

First, we will conduct a brief review of moments, spatial average pooling (SAP) and its relationship to convolution.

Given a random variable $X \in R$, the $i^{th}$ moment is defined as $E[X^i], i = {1, ..., n}$, and the $i^{th}$ central moment is defined as $E[(X - E[X])^i], i = {2, ..., n}$. When $i = 1$, the moment is the mean $\mu$ of $X$. When $i = 2$, the central moment is the variance $\sigma^2$ of $X$. When $i = 3$, the central moment is the unnormalized skewness of $X$, which is defined as $E[(\frac{X - E[X]}{\sigma})^3]$. When $i = 4$, the central moment is the unnormalized kurtosis of $X$, which is defined as $E[(\frac{X - E[X]}{\sigma})^4]$. In practice, higher order moments with $i > 4$ are less commonly used.

On the other hand, given a feature map $f \in R^{C \times H  \times W}$, the SAP is the operation of computing the $i = 1$ moment of values inside a pooling window to output the pooled feature map $f_{SAP} \in R^{C \times H' \times W'}$ (See Fig. 2). Identical to the standard convolution operation, the outputted $H'\times W'$ are determined by kernel size, stride, dilation and padding. 

In fact, for any pooling, the pooling window is the same as convolution windows given same settings. Besides, it can also be implemented as \textit{im2col}. The difference between pooling and convolution is after extracting windows to row vectors. In convolution, the extracted rows $r_i^T$ is dot produced with the flattened kernel. In SAP, the mean of $r_i^T$ is computed. This makes SAP a special form of convolution with box filter kernel. However, the spatial moment pooling-n in Section 3.2 can not be generalized by convolution when $n \ge 2$, since the $i \ge 2$ moments are non-linear and convolution is linear.

\subsection{Spatial Moment Pooling-n / SMP(n)}

\begin{figure}[htb]
    \includegraphics[width=8cm]{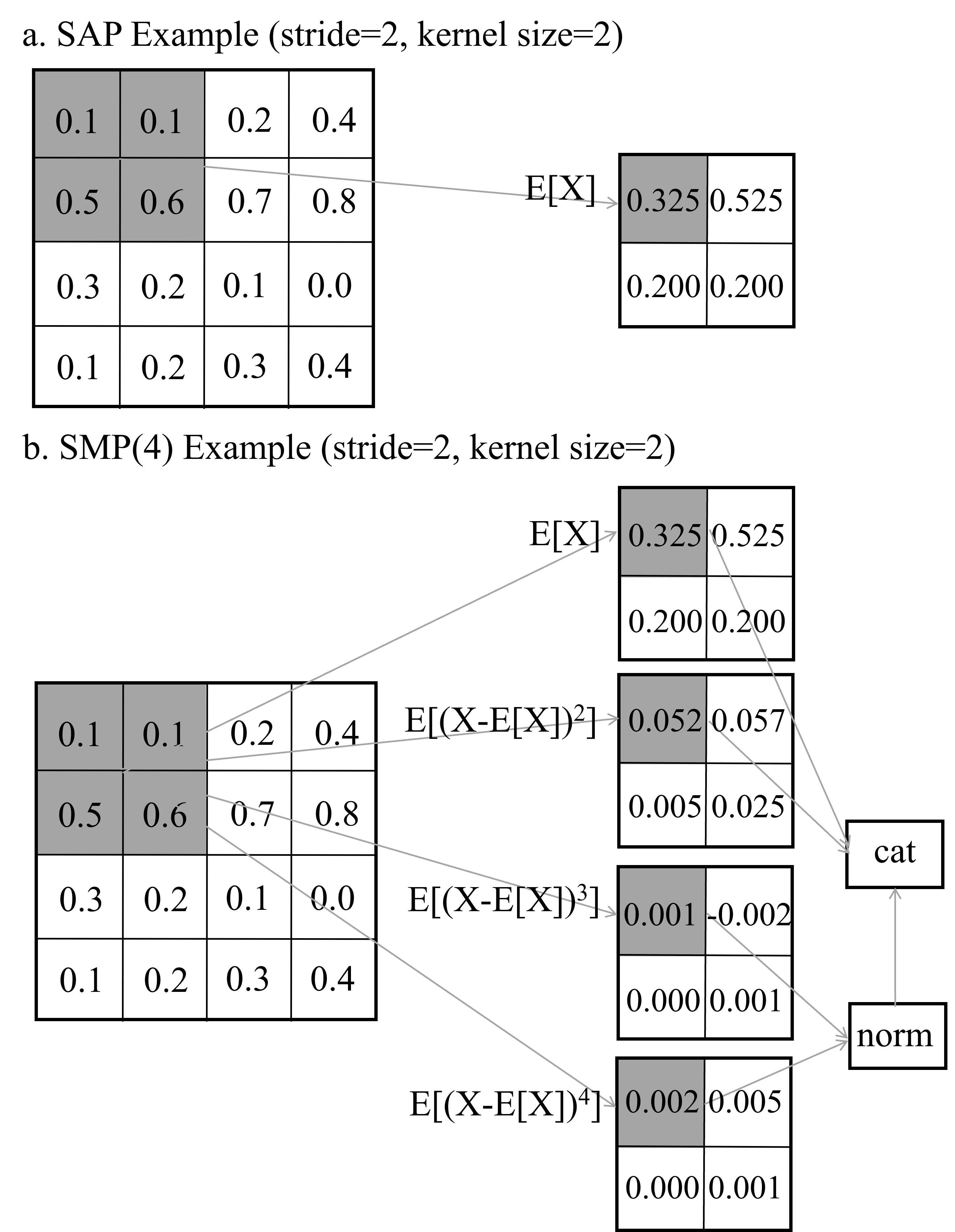}
    \label{fig:smp}
    \caption{a. spatial average pooling/SAP illustrated. b. spatial moment pooling-4/SMP(4) illustrated.}
\end{figure}

Similarly, we define the spatial moment pooling-n/SMP(n) as the operation of computing the $i = {1, ..., n}$ central moments of a pooling window, concatenating the moments in channel dimension and outputting the pooled feature map $f_{SMP} \in R^{nC \times H' \times W'}$ (See Fig. 2). The outputted spatial dimensions $H'\times W'$ are the same as SAP. However, the channel size increases from $C$ to $nC$. SMP(n) is a generalization of SAP. In fact, SMP(1) is exactly the same as SAP.

To intuitively show why SMP(n) might represent image quality better, we provide a toy example in Fig. 3. Two $3 \times 3$ feature maps with same mean are shown. One is checkerboard pattern, and the other is solid color. Despite the same means, the $i>1$ central moments differ.

\begin{figure}[htb]

\includegraphics[width=8cm]{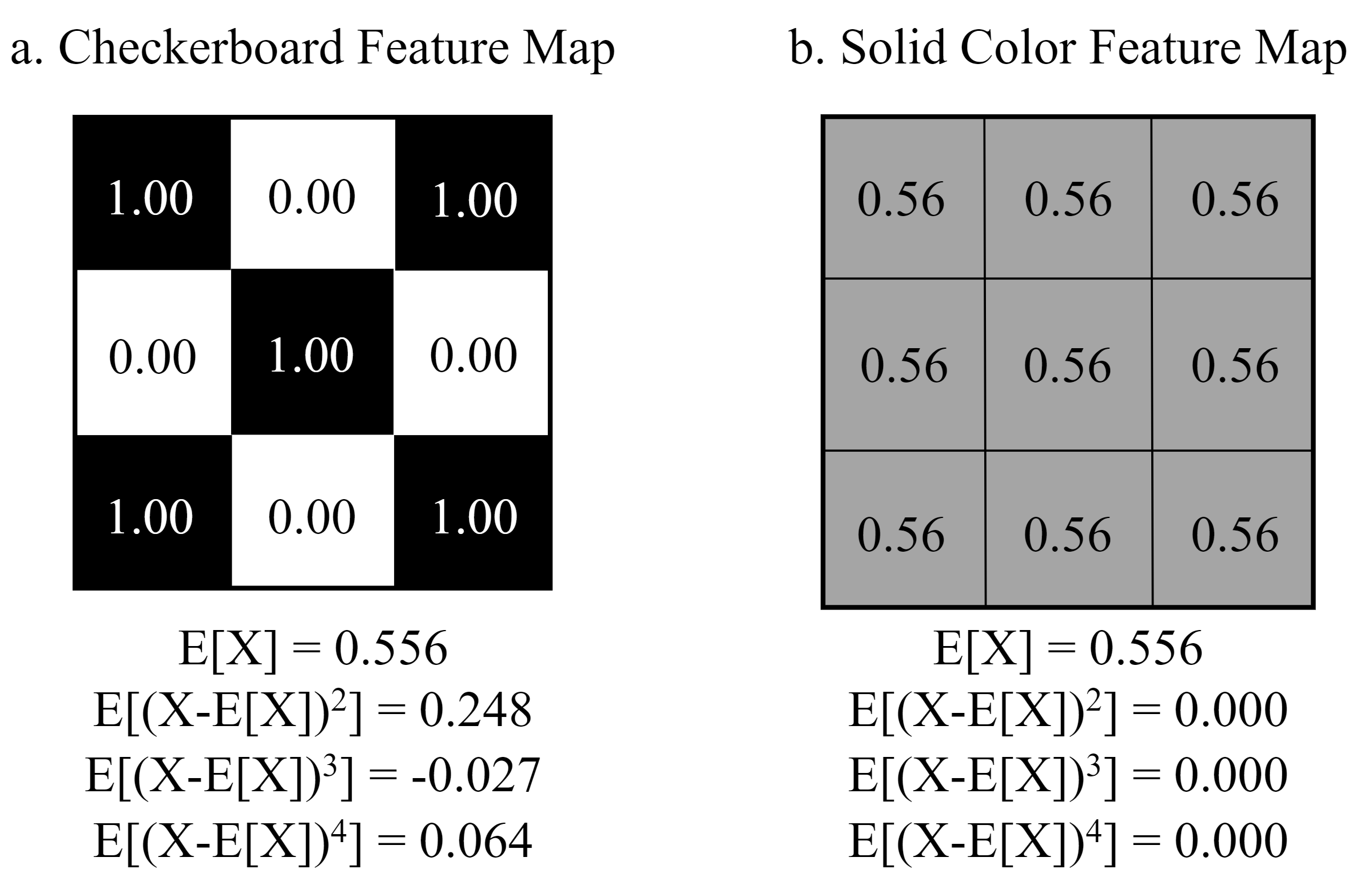}
\label{fig:example}
\caption{The $1^{st}$ moment, $2^{nd}, 3^{rd}, 4^{th}$ central moment of a $3 \times 3$ checkerboard feature map and solid color feature map.}

\end{figure}

\subsection{Numerical Difficulties and Normalization}
Simply elevating SMP(n) to $n \ge 3$ cases can cause severe difficulties when optimizing the network with back-propagation. As shown in Tab.~\ref{tab:ablation}, the training fails and produces NaN results without proper normalization. So we propose to add layer normalization after $i \ge 3$ central moments (See Fig. 2). The rationale of normalizing only $i \ge 3$ is: (1) $SMP(n), n \le 2$ works well without normalization. (2). only the statistical skewness and kurtosis are normalized with $\sigma$, the $\mu$ and $\sigma$ themselves are defined as unnormalized.

The reason why to choose layer normalization over others is purely empirical. The comparison of normalization methods is detailed in Section 4.2.2. In the following sections, SMP(n) with $n \ge 3$ implies that $i \ge 3$ moments are layer normalized  if no normalization method is specified.

\section{Experiments}
\subsection{Experiment Setup}

The SMP(n) is implemented in Pytorch 1.8, all the experiments are conducted on a computer equipped with Intel Xeon E5-2620 v4 and 8 Nivida TitanXp GPU.

Basically, there are two types of datasets for blind IQA. One is large scale, natural datasets without reference images (such as Koniq-10k \cite{hosu2020koniq}). The other is small scale synthetic datasets with reference images and distortion information (such as LIVE \cite{sheikh2006statistical}). Usually the methods designed for small dataset use reference and distortion as auxiliary task, which makes them not applicable to large datasets. Consequently, the sota methods for large and small dataset are different.

For large dataset, we choose Koniq-10k as dataset, and HyperIQA \cite{su2020blindly} as current CNN-based sota method. For small dataset we choose LIVE \cite{sheikh2006statistical} and CSIQ \cite{larson2010most} as dataset, and NemgIQA \cite{chen2021nested} as sota method. For ablation study we use Koniq-10k as dataset, and a variant of NIMA \cite{chadha2021deep}\cite{talebi2018nima} as base method. The dataset splits are the same as \cite{su2020blindly} and \cite{chen2021nested}.

\subsection{Ablation Study}
For baseline, we use an contemporary variant of classical NIMA \cite{talebi2018nima}. As proposed and described in \cite{chadha2021deep}, this variant incorporates the multi-scale spatial average feature maps. To simplify notations in the tables below, we denote this variant as NIMA. We only conduct experiments on VGG16 backbone, as the selection of CNN backbone selection is beyond scope of this paper.

\begin{table}[ht]
\centering
\caption{Ablation study results.}
\vspace{2mm}
\label{tab:ablation}
\begin{tabular}{lcc}
\toprule
                      & \multicolumn{2}{c}{Koniq-10k} \\ \cmidrule(l){2-3}
                      & SRCC         & PLCC          \\ \midrule
NIMA \cite{talebi2018nima}\cite{chadha2021deep} / NIMA-SMP(1) & 0.864         & 0.879         \\
NIMA-Vanilla-Large            & 0.864         & 0.874         \\
NIMA-SMP(2)           & 0.886         & 0.896         \\
NIMA-SMP(4)-w/o-Norm   & NaN           & NaN           \\
NIMA-SMP(4)-BatchNorm & 0.880         & 0.892         \\
NIMA-SMP(4)-MaxNorm   & 0.888         & 0.899         \\
NIMA-SMP(4)-LayerNorm & \textbf{0.890}         & \textbf{0.900}         \\ \bottomrule
\end{tabular}
\end{table}

\subsubsection{Effects of spatial moment pooling}
Tab.~\ref{tab:ablation} shows that NIMA-SMP(2) evidently improves both PLCC and SRCC over NIMA. NIMA-SMP(4)-LayerNorm outperforms SMP(2) despite the gain is not as significant as NIMA-SMP(2) over NIMA. Therefore, we stop at $n=4$ considering that the moments higher than $4^{th}$ are also uncommon in statistics.

\subsubsection{Effects of high moments normalization}
Simply expanding NIMA-SMP(2) to NIMA-SMP(4)-w/o-Norm brings optimization difficulties. After several iterations of training, the network produces constant results regardless of input images. And such constant output has $0$ variance, which further leads to NaN in PLCC and SRCC (See Tab.~\ref{tab:ablation}).

Although the NaNs can be eliminated by introducing batch normalization to $3^{rd}, 4^{th}$ moments, it also brings performance decay. The NIMA-SMP(4)-BatchNorm is outperformed by NIMA-SMP(2), this confirms the observations that batch normalization negatively effects scale sensitive low level computer vision tasks \cite{yu2018wide}.

Empirically, we find that max and layer normalization enable NIMA-SMP(4) to produce a reasonably superior performance over NIMA-SMP(2). Moreover, NIMA-GMP(4)-LayerNorm outperforms all other methods. 

\subsubsection{Effects of network parameter increment}
The SMP's increment of parameters is of no significance. For NIMA-SMP(4)-LayerNorm, the parameter is increased by $0.20\%$ compared with NIMA. And for NIMA-SMP(2), the parameter increase is only $0.068\%$. Moreover, the increment of MACs is also minimal. The base model NIMA has a MAC of 635.23G Mac (multiply-accumulate) when input image size is 3x1920x1080. Replacing last pooling with SMP-(2) brings 1.02M extra Macs. And replacing last pooling with SMP-(4) brings 3.03M extra Macs.

To fully verify the performance improvement comes from SMP instead of parameter increase, we build NIMA-Vanillia-Large with much wider and deeper regression head. NIMA-Vanillia-Large increases the parameter by as much as $14.74\%$. However, Tab.~\ref{tab:ablation} shows that the performance does not benefit from na\"ive model size increase.

\subsection{Results on Large Dataset}
\begin{table}[ht]
\centering
\caption{Results on Koniq-10K datasets. Blue and black numbers in bold represent the best and second best respectively. Reference methods' results are sourced from \cite{chen2021nested}\cite{ke2021musiq}.}
\vspace{2mm}
\label{tab:large}
\begin{tabular}{lcc}
\toprule
                & \multicolumn{2}{l}{Koniq-10K} \\ \cmidrule(l){2-3}
                & SRCC          & PLCC          \\ \midrule
\multicolumn{3}{l}{\textit{DCNN based methods}}                 \\
BIECON \cite{kim2016fully}      & 0.618         & 0.651         \\
PQR \cite{zeng2018blind}        & 0.880         & 0.884         \\
SFA  \cite{li2018has}           & 0.856         & 0.872         \\
NIMA \cite{talebi2018nima}      & 0.864         & 0.879         \\
DBCNN \cite{zhang2018blind}     & 0.875         & 0.884         \\
HyperIQA \cite{su2020blindly}   & 0.906         & 0.917         \\ \midrule
\multicolumn{3}{l}{\textit{DCNN + SMP based methods}}           \\
HyperIQA-SMP(2) & \textbf{0.909}         & \textbf{0.924}         \\ 
HyperIQA-SMP(4) & \textbf{0.909}         & 0.922         \\ \midrule
\multicolumn{3}{l}{\textit{Transformer based methods}}           \\
MUSIQ-Singlescale & 0.905         & 0.919         \\ 
MUSIQ-Multiscale & \textcolor{blue}{\textbf{0.916}} & \textcolor{blue}{\textbf{0.928}}   \\ \bottomrule

\end{tabular}
\end{table}

For HyperIQA on large dataset, we replace the global average pooling in local distortion aware modules. Moreover, we set the learning rate to $1 \times 10^{-5}$. The other settings are identical to original HyperIQA \cite{su2020blindly}. Results in Tab.~\ref{tab:large} show that such modification effectively improves HyperIQA, and a new sota of deep CNN based approaches is made. 

Notably, our HyperIQA-SMP(2) and SMP(4) are superior to the transformer with single scale image patch input. Although the multi-scale transformer outperforms SMP methods, it is not quite fair to draw comparison as MUSIQ-Multiscale requires multi-scale image inputs. We also note that it is possible to train HyperIQA-SMP(n) with multi-scale image crops, but the effects of single-scale and multi-scale input is beyond the scope of this paper.

\subsection{Results on Small Dataset}
\begin{table}[h]
\centering
\caption{Results on LIVE and CSIQ datasets. Blue and black numbers in bold represent the best and second best respectively. Reference methods' results are sourced from \cite{chen2021nested}\cite{ke2021musiq}.}
\vspace{2mm}
\label{tab:my-table}
\begin{tabular}{lcccc}
\toprule
               & \multicolumn{2}{c}{LIVE} & \multicolumn{2}{c}{CSIQ} \\ \cmidrule(l){2-3} \cmidrule(l){4-5} 
               & SRCC       & PLCC       & SRCC       & PLCC       \\ \midrule
\multicolumn{5}{l}{\textit{DCNN based methods}}               \\
BIECON \cite{kim2016fully}       & 0.963       & 0.965      & 0.837       & 0.858      \\
PQR \cite{zeng2018blind}         & 0.965       & 0.971      & 0.873       & 0.901      \\
SGDNet \cite{yang2019sgdnet}     & 0.961       & 0.964      & 0.878       & 0.909      \\
HyperIQA \cite{su2020blindly}    & 0.961       & 0.963      & 0.914       & 0.927      \\
NemgIQA \cite{chen2021nested}    & 0.971       & 0.975      & 0.923       & 0.934      \\ \midrule
\multicolumn{5}{l}{\textit{DCNN + SMP based methods}}               \\
NemgIQA-SMP(1)    & 0.972       & 0.978      & 0.921       & 0.937      \\
NemgIQA-SMP(2) & \textbf{0.973} & \textbf{0.980} & \textcolor{blue}{\textbf{0.929}} & \textbf{0.942} \\
NemgIQA-SMP(4) & \textcolor{blue}{\textbf{0.976}} & \textcolor{blue}{\textbf{0.982}} & \textbf{0.924} & \textcolor{blue}{\textbf{0.945}} \\ \bottomrule
\end{tabular}
\end{table}

For NemgIQA on small dataset, it is not possible to directly replace last global average pooling with SMP. Unlike other works with regression performed after pooling, NemgIQA uses a single channel output point wise convolution as regression head followed by pooling. To make SMP usable, we swap the order of pooling and regression back. To be speficic, we change last layers to: point wise regression with output channels 4, pooling, and an affine layer to produce score. The other settings are the same as original NemgIQA \cite{chen2021nested}.

Due to such modifications NemgIQA-SMP(1) is not strictly equivalent to original NemgIQA. However, experimental results show that this does not effects its performance. Moreover, the NemgIQA-SMP(2) and NemgIQA-SMP(4) significantly improve PLCC and SRCC compared to both original NemgIQA and NemgIQA-SMP(1), and achieve sota performance on LIVE and CSIQ datasets.

\section{Conclusion}

In this paper, we extend spatial average pooling (SAP) into spatial moment pooling (SMP) by adding higher order moments. Moreover, we propose an optimization friendly normalization trick that makes training of network with SMP stable. Experimental results show that replacing SAP with SMP significantly improves deep CNN-based blind IQA approaches.

\bibliographystyle{IEEEbib}
\bibliography{strings,refs}

\end{document}